\documentclass[a4paper,fleqn]{cas-dc}

\usepackage{natbib}
\usepackage{graphicx}%
\usepackage{multirow}%
\usepackage{amsmath,amssymb,amsfonts}%
\usepackage{amsthm}%
\usepackage{mathrsfs}%
\usepackage[title]{appendix}%
\usepackage{xcolor}%
\usepackage{textcomp}%
\usepackage{manyfoot}%
\usepackage{booktabs}%
\usepackage{algorithm}%
\usepackage{algorithmicx}%
\usepackage{algpseudocode}%
\usepackage{listings}%
\usepackage{tikz}
\usepackage{pifont}
\usepackage{subcaption}
\usepackage{tabularx}
\usepackage{makecell}

\newcommand{\googletranslate}{%
    \begin{tikzpicture}[scale=0.05, baseline=-3pt]
          \node {\includegraphics[height=10pt]{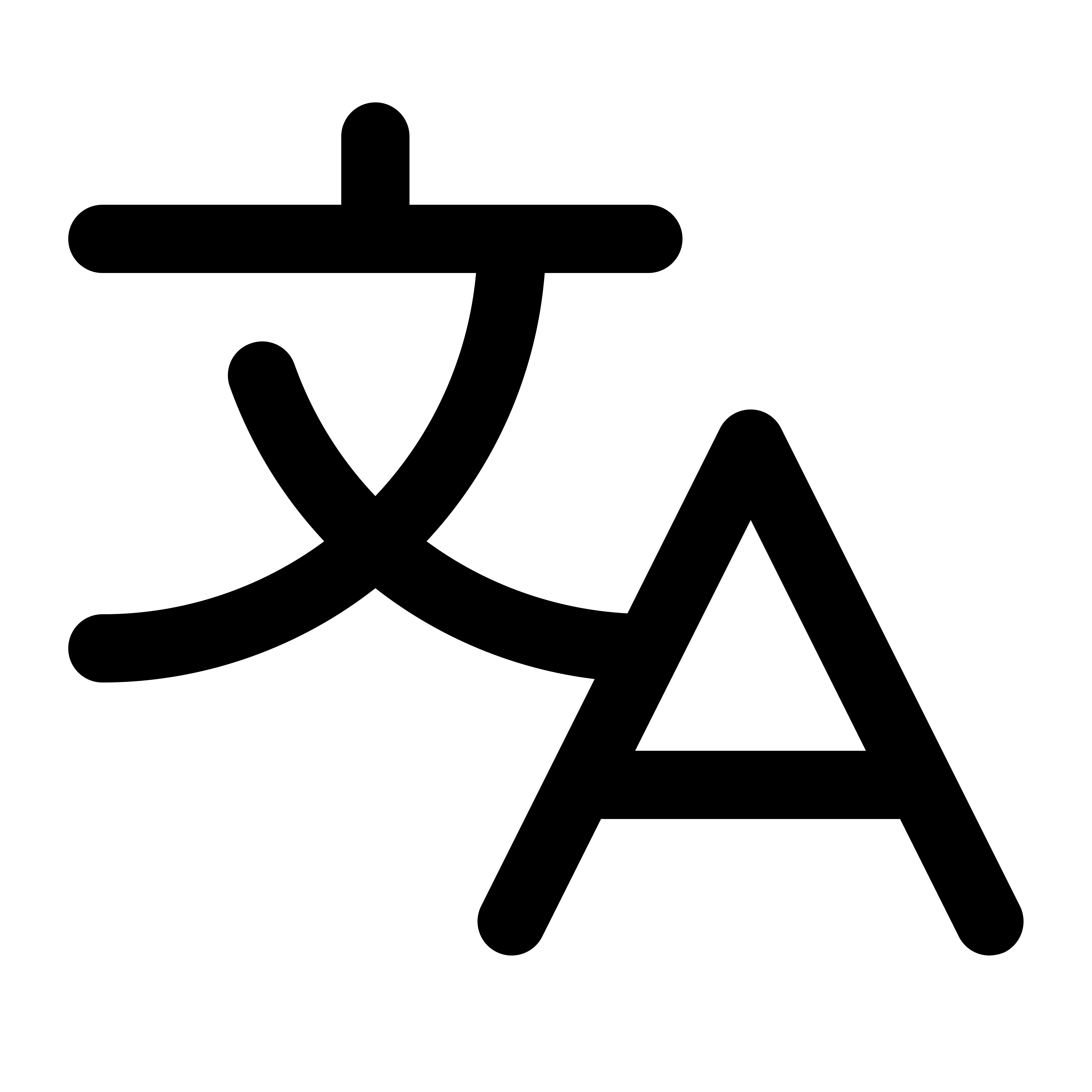}};
    \end{tikzpicture}%
}

\newcommand{\correlation}{ $\rho$ }

\newcommand{\literature}{%
    \begin{tikzpicture}[scale=0.05, baseline=-3pt]
          \node {\includegraphics[height=10pt]{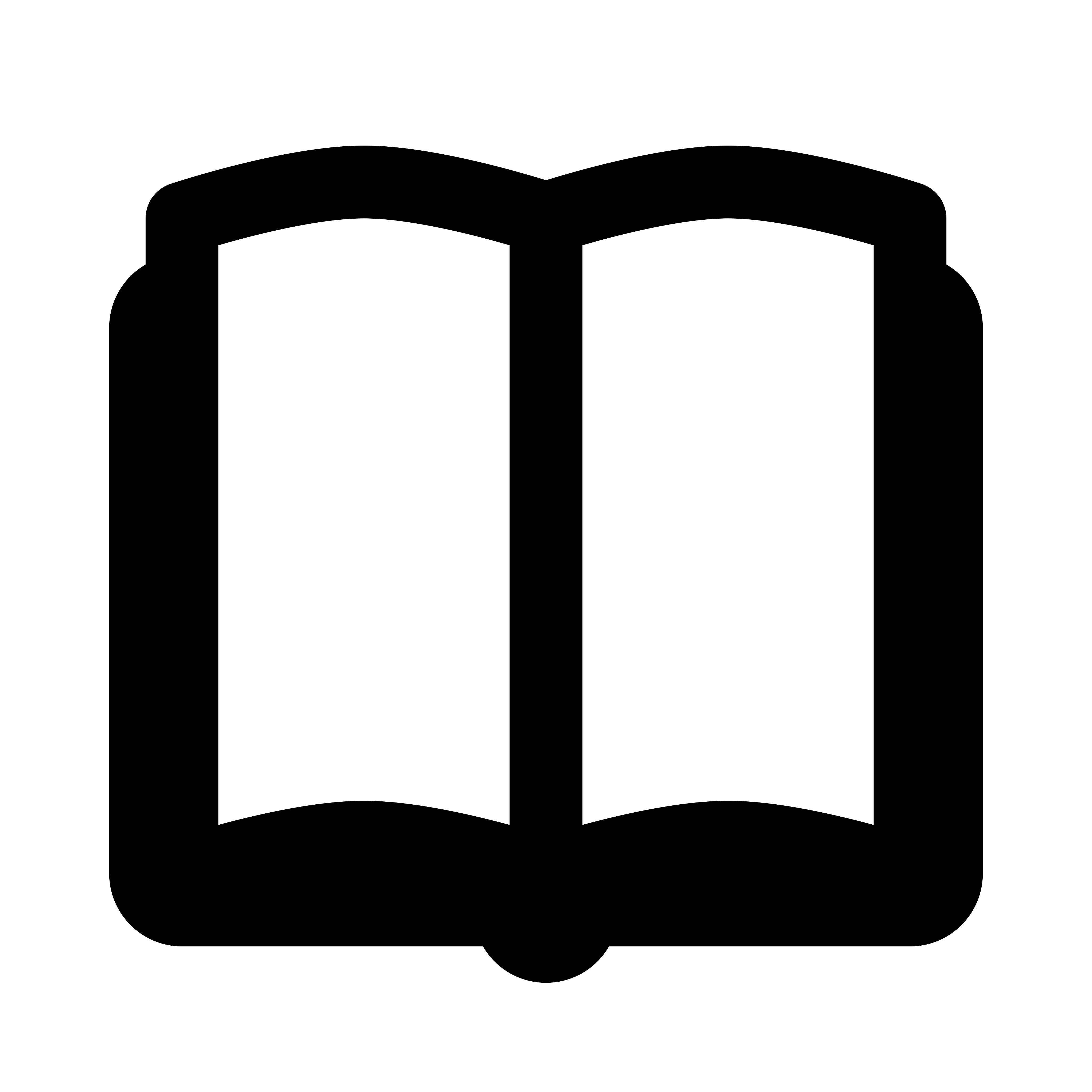}};
    \end{tikzpicture}%
}

\newcommand{\heuristic}{%
    \begin{tikzpicture}[scale=0.05, baseline=-3pt]
          \node {\includegraphics[height=10pt]{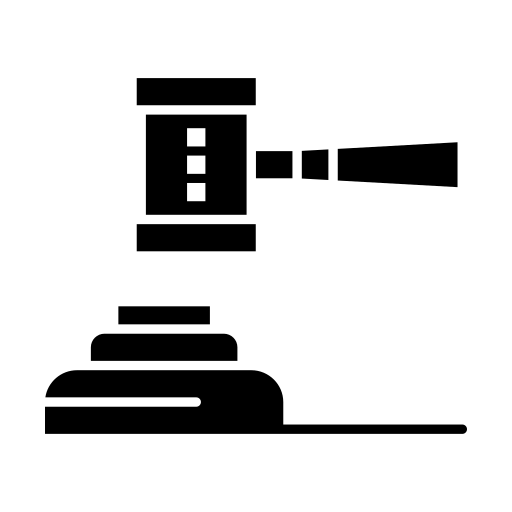}};
    \end{tikzpicture}%
}

\begin{document}

\title[mode = title]{Is Our Chatbot Telling Lies? Assessing Correctness of an LLM-based Dutch Support Chatbot}
\shorttitle{Is Our Chatbot Telling Lies?}
\shortauthors{H. Lassche et~al.}

\author[1]{Herman Lassche}[orcid=0009-0005-8764-4988]
\cormark[1]
\fnmark[1]
\ead{herman.lassche@afas.nl}
\author[1]{Michiel Overeem}[orcid=0000-0003-4807-4124]
\ead{michiel.overeem@afas.nl}
\author[2]{Ayushi Rastogi}[orcid=0000-0002-0939-6887]
\ead{a.rastogi@rug.nl}

\affiliation[1]{organization={Product Development, AFAS Software},
                city={Leusden},
                country={The Netherlands}}
\affiliation[2]{organization={Faculty of Science and Engineering, University of Groningen},
                city={Groningen},
                country={The Netherlands}}
                
\cortext[cor1]{Corresponding author}
\fntext[fn1]{During the research, the first author was a student at the University of Groningen and affiliated with AFAS as an intern.}

\begin{abstract}
Companies support their customers using live chats and chatbots to gain their loyalty. AFAS is a Dutch company aiming to leverage the opportunity large language models (LLMs) offer to answer customer queries with minimal to no input from its customer support team. Adding to its complexity, it is unclear what makes a response correct, and that too in Dutch. Further, with minimal data available for training, the challenge is to identify whether an answer generated by a large language model is correct and do it on the fly. 

This study is the first to define the correctness of a response based on how the support team at AFAS makes decisions. It leverages literature on natural language generation and automated answer grading systems to automate the decision-making of the customer support team. We investigated questions requiring a binary response (e.g., Would it be possible to adjust tax rates manually?) or instructions (e.g., How would I adjust tax rate manually?) to test how close our automated approach reaches support rating. Our approach can identify wrong messages in 55\% of the cases. This work demonstrates the potential for automatically assessing when our chatbot may provide incorrect or misleading answers. Specifically, we contribute (1) a definition and metrics for assessing correctness, and (2) suggestions to improve correctness with respect to regional language and question type.
\end{abstract}


\maketitle

\section{Introduction}
Companies value their customers~\citep{VirtualAgents} and strive to create a great customer experience \citep{CustomerExperience}. Customers, in turn, assess a company on its core business and customer service, which influences their trust, loyalty, and satisfaction~\citep{CustomerServiceBTB}. Today, the most popular way to assist customers online is via chatbots and live chats~\citep{ChatTools, BusyHelp, VirtualAgents}. Real-time communication in live chats means quick answers to questions \citep{BusyHelp, VirtualAgents}, which helps build loyalty \citep{ChatTools} and encourages customers to return when they need assistance \citep{BusyHelp}.

\begin{figure}
\centering
\includegraphics[width=\linewidth]{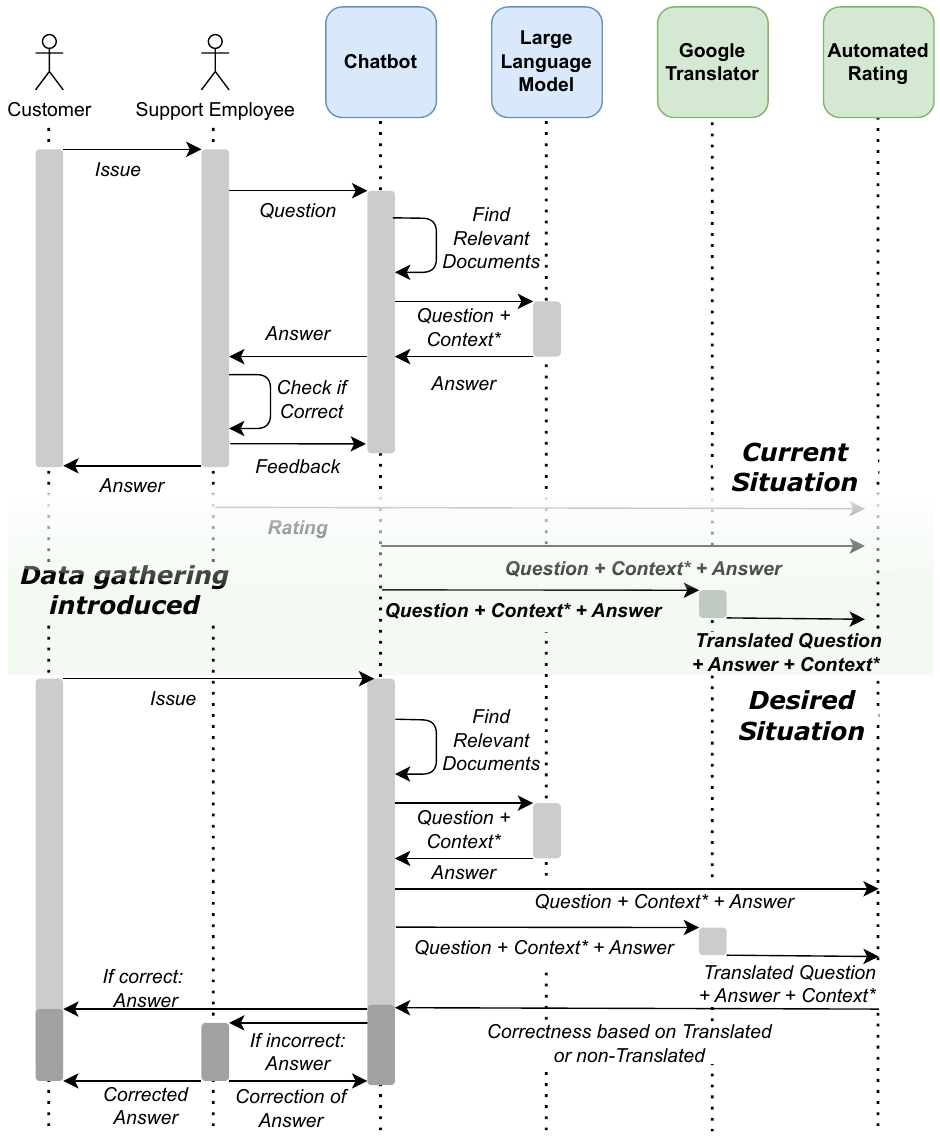}
\caption{Shows the current and desired flow for handling customer queries. In the current workflow, the support team is an intermediate for providing context* comprising of relevant documents and instructions for the large language model and later assessing the response (see 'Chatbot' and 'Large Language Model'). Using parts 'Google Translator' and 'Automated Rating', we envision replacing human feedback with automated ratings}
\label{fig:chatFlow}
\end{figure}

With recent advancements in Large Language Models (LLM) that enable natural and chat-like communication \citep{EvolutionLLM, AdvancesinAI, GPT-4ChatBot}, AFAS sees an opportunity to provide live support. The upper half of Figure \ref{fig:chatFlow} represents the current situation. When a customer raises an issue, an employee forwards the question to the chatbot along with relevant documents and instructions, also called a system prompt. When the LLMs generates an answer based on the information provided (details in \ref{sec:industrialsetting}), the support employee checks the answer and forwards it to the customer if correct. 

Moving forward, we envision minimizing the validation by the support team, creating room for the support team to handle complex issues, and improving customer experience through near-real-time response. We aim to create an automated solution to identify lies our LLM-based support chatbot tells, indicative of the quality of the response, which is crucial for customer satisfaction. Further, the solution should be in Dutch to cater to the Dutch audience.  

Our first challenge in building an automated solution is understanding what makes a response right. To solicit an answer to this question, the first author shadowed a support staff for a day to observe and interactively understand what makes a response right. Combined with the literature search and analysis of reasons for rejecting chatbot responses, this showed that the first step to the `right’ answer is \emph{correctness}, characterized in terms of \emph{relatedness, completeness, and truthfulness}. With much research focusing on relatedness~\citep{Merdivan-Singh-Hanke-Kropf-Holzinger-Geist-2020, zhang2020bertscore}, this study focuses on truthfulness. To assess our approach, in the first round we gathered a data of 79 posts which we used for training. At a later point, we collected data from 154 posts for testing. The limited data size characterize our study.

Since the training data is scarce, it is not possible to train a model using reference answers as has been widely seen in literature~\citep{papineni-etal-2002-bleu, banerjee2023benchmarking, zhang2020bertscore, Kumar-Aggarwal-Mahata-Shah-Kumaraguru-Zimmermann-2019, roy2016iterative}. As an alternative, we model how the support team makes decisions and derive heuristics. To measure these heuristics, we take inspiration from Natural Language Generation \citep{zhang2020bertscore, banerjee2023benchmarking, Roychowdhury-Alvarez-Moore-Krema-Gelpi-Agrawal-Rodríguez-Rodríguez-Cabrejas-Serrano-et} and Automated Answer Grading literature for metrics \citep{roy2016iterative, Kumar-Aggarwal-Mahata-Shah-Kumaraguru-Zimmermann-2019, Jamil-Hameed-2023, P-B-2022, Vij-Tayal-Jain-2019}. In the process, we note that the choice of heuristics varies with the type of question asked. For example, heuristics for assessing the correctness of a yes/no answer are different from the heuristics for a question that solicits instructions. 

Our resulting model assesses the correctness of yes/no questions and questions requiring instructions to show that our model can detect a very inaccurate response with 55\% accuracy. Notably, the overall accuracy is better for translated text in English than in Dutch. Further, we observed a 0.3 correlation of our score with human evaluation for Dutch text and 0.37 for the translated text in English, both of which is higher than the 0.13 reported by \cite{Mehri-Eskenazi-2020} in their study on generic conversations.

Further, our study contributes by providing
\begin{itemize}
\item a working definition and metrics to assess the correctness of the responses generated by LLMs
\item suggestions to improve the correctness for regional language and type of question
\end{itemize}

The proposed definition provides a structured foundation for feedback. We decomposed the definition of correctness into smaller, well-defined definitions. Making chatbot evaluation more concrete and manageable. Our methodology further demonstrates how this definition can be utilized in an empirical evaluation. Moreover, the proposed metric and features may be directly applied or serve as foundation for developing customized evaluation metrics for LLMs. Through our contributions, we illustrate how software companies developing chatbots can implement guardrails for their systems. These metrics will help the system to prevent the bot from telling lies and thus improving the quality of the chatbot. Finally, our recommendations for improving correctness in regional languages can serve to further advance the evaluation of chatbots.

\section{Industrial Setting}\label{sec:industrialsetting}
AFAS is a software company that specializes in automating business processes through their ERP system.
Its headquarter resides in The Netherlands, but AFAS has offices in Belgium and on the Caribbean too.
The software product is provided as a service to thousands of organizations. 
In 2023 more than 3 million users utilized their software, leading to almost 112,000 support inquiries.\footnote{See annual report: https://jaarverslag.afas.nl/2023} 
These support queries are handled by a support team of 70 people who dedicate their time to this task.
Since the support team receives nearly one query every minute about their software, saving time on even a subset of the queries will be helpful.

To grow its user base without proportionally growing the support team, AFAS started to develop a chatbot that automates the answering of support queries.
This chatbot is developed by a team consisting of four developers.
They use internal developed frameworks for both back-end processing and front-end rendering.

Using LLMs and grounding the prompts with relevant documentation, the AFAS development team hopes to unburden the support team by not only providing them with a possible answer, but also showing them which documentation is relevant.
The AFAS chatbot uses Retrieval Augmented Generation (or RAG) to improve the performance of the language model. RAG comprises four parts \citep{yu2024evaluation}: 

\begin{itemize}
    \item Indexing | Creates an index of all documents containing relevant information for a user. This step is done before the chatbot serves any answers. At the time of writing, AFAS indexes help documents. 
    \item Search | Relevant documents are retrieved based on their similarity to the user message using embedding similarity \citep{RAGEmbedding} and keyword-matching techniques \citep{RAGTFIDF, RAGBM25}. Figure \ref{fig:chatFlow} shows this step as \textit{Find Relevant Documents}.
    \item Prompting | This step combines the user message, relevant documents, and system prompt as a single message. The system prompt includes instructions and basic information, such as 'be friendly' and contact information. Since the chatbot relies on an LLM that is not fine-tuned on company data but is a generally trained model, the LLM requires necessary information to provide relevant responses. Relevant documents provide this information to the LLM. This is shown as \textit{Question + Context} from Chatbot to LLM in Figure \ref{fig:chatFlow}.    
    \item Inference | The question + context is used to prompt an LLM, and the generated response is shown to the support team. This is depicted as the \textit{answer} from the LLM to the chatbot in Figure \ref{fig:chatFlow}.
\end{itemize}

We envision the chatbot to handle all kinds of questions, considering the unique jargon of the company/industry and the fact that it mainly serves Dutch users. 
Here, reference answers may not help since they are sparse and do not ensure a right response to unseen use cases. 
Unseen questions are expected, as we consider a user-driven chatbot \citep{10.1007/978-3-030-17705-8_13}. 
Users can ask a wide variety of questions, which means the metric must be prepared to handle unseen questions. 
Given these constraints, there is a need for a generic definition of what makes a right answer and how to measure it. 

\section{What makes a right answer?}\label{sec:whatsright}
There are two ways to assess whether a chatbot gives right answers: turn- and dialogue-level metrics. Turn-level metrics rate a single message-answer pair \citep{zhang2020bertscore, Singh-Suraksha-Nirmala-2021, Yan-Song-Wu-2016, banerjee2023benchmarking, Das-Verma-2020, Phy-Zhao-Aizawa-2020, Tao-Mou-Zhao-Yan-2018, Gupta-Rajasekar-Patel-Kulkarni-Sunell-Kim-Ganapathy-Trivedi-2022, Roychowdhury-Alvarez-Moore-Krema-Gelpi-Agrawal-Rodríguez-Rodríguez-Cabrejas-Serrano-et, Mehri-Eskenazi-2020}. In contrast, dialogue-level metrics rate the full dialogue, including all message-answer pairs \citep{Yeh-Eskenazi-Mehri-2021, Huang-Ye-Qin-Lin-Liang-2020, Pang-Nijkamp-Han-Zhou-Liu-Tu-2020, deriu2020spot, Lowe-Noseworthy-Serban-Angelard}. Our objective is to assess the correctness of each answer and, hence, turn-level. In addition, the goal of the chatbot is to provide a correct answer in its first response, without requiring any further interaction.

Correctness differs from the commonly known term hallucination. As hallucinated answers (not grounded in the context) can be correct \citep{Ji_2023, ji2023mitigatinghallucinationlargelanguage}. Conversely, non-hallucinated answers can be incorrect if important content is absent \citep{Ji_2023}.

To the best of our knowledge, no prior work defines correctness or not clear enough \citep{Mehri-Eskenazi-2020, Ji_2023} for measurement and validation. Therefore, our first objective was to define correctness. To define correctness, we followed a two-pronged approach. First, we looked at 500 chatbot responses for which the support team provided a decision: accept or reject and a short justification for rejection. Further, the first author shadowed \citep{Shadowing} an experienced support employee.

Based on shadowing, discussion, and analysis of rejection reports, the three most common mistakes and, hence, requirements for correctness stood out. 
They are as defined in the \cite{Oxford}: 

\begin{center}
    {{Truthfulness}}\\
    \textit{"The quality of only saying what is true"},
\end{center}
\begin{center}
    {{Relatedness}}\\
    \textit{"A close connection with the subject you are discussing or the situation you are in"},
\end{center}
\begin{center}
    {{Completeness}}\\
    \textit{"The fact of including all the parts, etc. that are necessary; the fact of being whole"}.
\end{center}

The annotations often clearly reflected the three requirements. For instance, several comments explicitly indicated issues with completeness, such as: “Incomplete, Jonas (the chatbot) should also provide instructions on how to structure the tasks to proceed earlier.”

A response is considered correct if it contains only true information, as supported by multiple research studies \citep{li-etal-2018-ensure,wang-etal-2021-sketch,ji2023mitigatinghallucinationlargelanguage} (truthfulness), is related to the situation and question (relatedness), and comprises all relevant information and solutions (completeness). Later, we designed a plugin soliciting support team responses to understand which requirements the chatbots fall short of, as visualized in Figure \ref{fig:chatFlow}. 

Returning to the literature, we observed that relatedness is well-researched \citep{zhang2020bertscore, Singh-Suraksha-Nirmala-2021, Yan-Song-Wu-2016}. However, truthfulness and completeness of generated answers are not. Of the remaining two, we study truthfulness since if an answer is not true, completeness would not matter. For an incomplete answer, the customer can ask follow-up questions, but if untrue information is presented to the customer, it can cause harm to the customer and the company. In the future, a combination of the above three dimensions can be used to measure correctness. The rest of the paper measures truthfulness and assesses it with respect to the manually rated ground truth. 

\section{Methodology}
We combined qualitative and quantitative approaches to define and evaluate answer correctness in the chatbot. First, we analyzed existing feedback, manual annotations, and shadowed a support employee to understand how correctness is assessed in practice (Section \ref{sec:defining}). Based on these insights and literature review, we defined requirements for a correct answer: completeness, relatedness, and truthfulness. We then implemented an extra annotation plugin to collect structured feedback from the support team along these requirements (Section \ref{sec:gathering}) and focused on truthfulness for automated evaluation. Using the annotated data, we built a decision tree to model the support team’s assessment process (Section \ref{sec:construct}). This led to the identification of message types (Section \ref{sec:types}), and we derived heuristics describing what makes an answer true (Section \ref{sec:identified}). These heuristics guided the selection of literature-based metrics to create automated features (Section \ref{sec:automated}). The automated features were combined into a truthfulness score (Section \ref{sec:features}) and validated on a test set (Section \ref{sec:evaluation}). Additionally, since most related research and tools are developed for English, we include a side experiment to explore whether using English translations of Dutch responses improves performance.

For ease of reading, you will also find parts of the methodology in Sections V and VI. In the following subsections, we describe the data we collected for analysis and training. Next, we construct a decision tree to capture how the support team makes decisions. This representation is closer to how the support team thinks and is hard to translate to metrics. Therefore, we introduce an intermediate step to identify heuristics from the decision tree. At this stage, we observed that not all heuristics are relevant for all message types, and therefore, we characterize message types and the heuristics that apply to each message type. In Section V, we searched the literature for metrics that likely represent the heuristics and carefully selected a subset for modeling. Finally, in Section VI, we assess the scores derived from the model with respect to the manually annotated rating from the support team collected in data collection. 

\subsection{Defining a right answer}\label{sec:defining}
To define what makes an answer right (Section \ref{sec:whatsright}), we began by conducting a literature search to identify existing definitions of right chatbot answers. In parallel, we analyzed annotations of the support team of AFAS. These annotations were gathered the months before we started our research. Each response generated by the chatbot could be either accepted or rejected by employees, with a short justification provided for each rejection. In total we were able to collect approximately 500 annotations, providing a rich source of quantitative feedback.

To complement the quantitative analyses, and gain a deeper insight in how support employees assess the rightness of answers, we collected qualitative data through shadowing \citep{Shadowing}. The first author shadowed an experienced support employee throughout one workday and interactively discussed the decision-making. The referenced employee has 12 years of experience in various product support areas of the software, has been using chatbot since its launch, and has received relevant AI training, including a course on using ChatGPT. During the discussions, the employee discussed special cases where the chatbot did not meet expectations and what modifications were required to ensure the answer was correct before presenting it to the customer.

The definition of correctness and its requirements were derived by integrating insights from three sources: (1) existing literature and quality metrics, (2) quantitative feedback from 500 chatbot evaluations, and (3) qualitative observations from employee shadowing. We first conducted open coding of the annotations to identify recurrent themes explaining why answers were accepted or rejected. These themes were then compared with concepts identified in prior literature. Through an iterative process of coding, discussion, comparison with existing frameworks, and clustering, we grouped the observed themes into three main requirements: truthfulness, relatedness and completeness. Which together capture the key requirements of a correct chatbot answer. This approach ensured that the resulting definition was both theoretically grounded and empirically validated within the real-world context of customer support.

\subsection{Data Gathering}\label{sec:gathering}
Currently, the developed chatbot of AFAS is utilized by the support team as an assistant. They use the bot to answer questions they have themselves or to answer the questions of a customer. Based on their expert knowledge, they rate the answer of the chatbot for truthfulness. They are encouraged to rate messages, but may choose which messages to rate by themselves. Consequently only a few messages are rated each day, this challenge is posed as we work with a real-case company scenario. The rating is along a Likert scale, which ranges from (1) very untrue to (5) very true. Likert scales are commonly used in similar research by \cite{jamanetworkopen.2023.36483}. While the size of the scale is a topic of debate, a 5-point Likert scale is most commonly used \citep{Borsci_Malizia_Schmettow_van, vanderLee, Voigt_2021}. During the research, the team continued to use the chatbot, which meant that new feedback was continually being received.

During the study, the plugin to rate truthfulness is implemented. A few weeks after the rating option is introduced, the data is extracted from the system to form an analysis set, consisting of 79 samples. A few weeks later, a test set is extracted, containing 154 message-answer pairs. For each rated message-answer pair, the context (relevant documents + system prompt) is gathered as well. Finally, to test whether English text performs better than Dutch, we translate the data using Google Translator\footnote{https://pypi.org/project/deep-translator/}. Since our raters frequently rate an answer as either fully true or fully untrue, the dataset becomes imbalanced, see Table \ref{tab:numberMessages}. Consequently, we focus less on overall accuracy and more on the accuracy of detecting 1-star and 5-star rated messages. These messages have a larger number and are of greater interest to the company. Since fully untrue answers must not be sent to users, and fully true answers can be sent without human intervention.

\begin{table}[htbp]
    \centering
    \begin{tabular}{cccccc}
        \hline
         & \bf 1 & \bf 2 & \bf 3 & \bf 4 & \bf 5 \\
        \hline
        Train set & 22 & 2 & 8 & 6 & 41 \\
        Test set & 32 & 5 & 19 & 16 & 82 \\
        \hline\\
    \end{tabular}\\
\caption{Counts of star ratings (1–5), assigned by the support team to reflect message truthfulness, in the training and test datasets.}
\label{tab:numberMessages}
\end{table}

\subsection{Constructing the Tree}\label{sec:construct}
To ensure that our metric correlates with human ratings, it is essential to understand how a human determines their rating. In order to represent the thought process of human annotators, we sought a model that could be easily visualized and understood. Furthermore, we sought a method to systematically organize the analysis and maintain a record of the observed characteristics. We opted for a decision tree to depict the mental process, with each node symbolizing the subconscious decisions made by the annotator. We use a manual created tree, as our goal is to mimic the human workflow. If we use automated decision tree builders like Random Forests \citep{RandomForest} or C4.5 \citep{Quinlan}, they would create their own reflection and would not reflect the human workflow.

In relevant literature, automated decision-makers utilize features that range from simple and syntactic to complex and semantic ones. Simple features might verify the presence of keywords \citep{Kumar-Aggarwal-Mahata-Shah-Kumaraguru-Zimmermann-2019, Jamil-Hameed-2023, Roychowdhury-Alvarez-Moore-Krema-Gelpi-Agrawal-Rodríguez-Rodríguez-Cabrejas-Serrano-et}, while more complex features involve evaluating the relevance \citep{Merdivan-Singh-Hanke-Kropf-Holzinger-Geist-2020, zhang2020bertscore} and meaning of a sentence \citep{Vij-Tayal-Jain-2019, leacock1998combining, resnik1995using}. The decision tree is build up based on this hierarchy, starting with syntactic checks and progressing to nuanced semantic evaluations if the syntactic checks hold. With this approach, the construction and human evaluation process becomes efficient, as it ensures that complex evaluations are not always necessary to be carried out.

The tree is initially constructed by the first author through iterative testing and modification using message-answer pairs. For the initial message-answer pair, the author assessed what makes the answer untrue and added a node describing the mistake. Then, another message-answer pair was evaluated to check if the decision tree properly rejects the answer. If not, a node is added or changed. This procedure is executed for all message-answer pairs and consistently cross-checked with the pairs that have already been evaluated. 

While the first version of the decision tree was proposed by the first author, it evolved iteratively to incorporate AFAS-specific context and maintain simplicity. Each of the other two authors contributed this perspective during the iterative process, and the AI development team at AFAS also provided input through discussions on the tree’s construction. Thus, although the tree was primarily built by the first author, it was developed in close consultation with others.

\begin{figure}
    \centering
    \includegraphics[width=\linewidth]{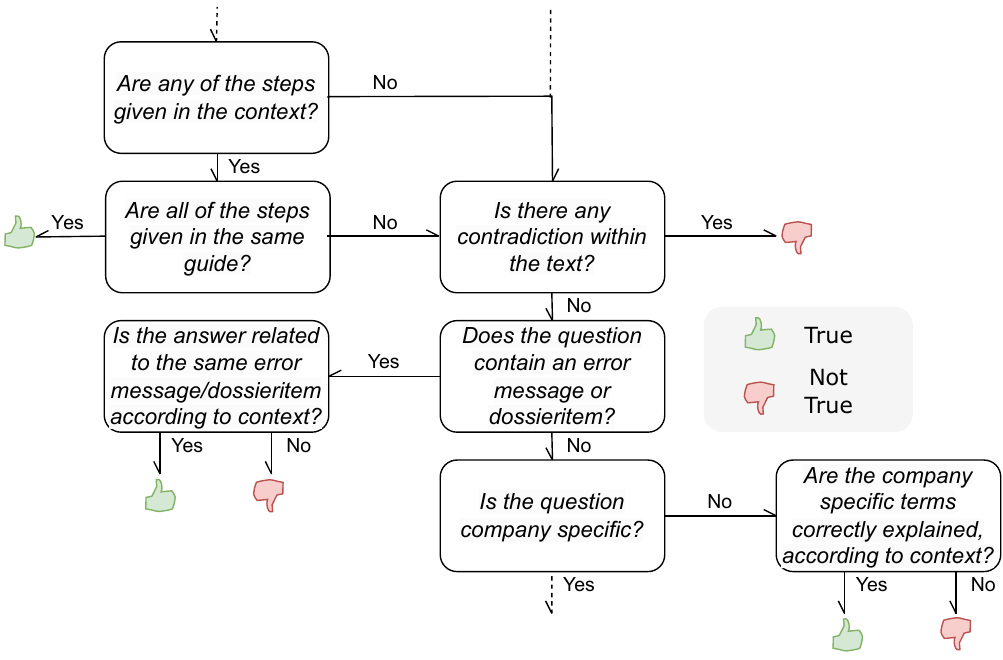}
    \caption{A snippet of the decision tree, indicating whether an answer would be true or not}
    \label{fig:partTree}
\end{figure}

As the decision tree is constructed manually, of course the representation differs if created by another researcher. However, multiple employees of the company confirm that the created decision tree correctly reflects the annotation process. Even when multiple people are involved, the decision tree reflects their interpretation of the mental model, which may differ from the actual mental model. However, by involving several contributors in the process, we believe we reduced this threat and arrived at a more generic and robust mental model. The final version of the decision tree can then be used to determine whether an answer is truthful. Although the decision tree does not make perfect decisions, it effectively visualizes the mental process of an annotator. A part of the tree is shown in Figure \ref{fig:partTree}, and the full tree is included in the replication package \citep{Replication}.

The first author has no knowledge about the product during the construction of the tree and has relied on the evaluation in the same context as our metric and the LLM. This way, the knowledge of the author is comparable to the knowledge of the LLM and the metric. It is crucial, as the metric, like the author, should be able to calculate a score without a reference answer and, therefore, without relying on external knowledge from such a reference. This is possible since the LLM is not trained, as far as we know on the company documentation. All info in the answer about the product should be
contained in the context; otherwise, the LLM will just come up with info on its own. If not in the context, the LLM is
assumed to be hallucinating \citep{Roychowdhury-Alvarez-Moore-Krema-Gelpi-Agrawal-Rodríguez-Rodríguez-Cabrejas-Serrano-et}.

The decision tree can be utilized by a human evaluator, as demonstrated in the following example. Referring to a snippet of the decision tree in Figure \ref{fig:partTree}, we examine the question: *I got error 404, what does it mean?* and the answer: *It means that you are not allowed to see the page.* We begin at the top right of the tree and determine that there is no contradiction in the answer. Next, we assess if the answer contains an error. Since it does, we only need to verify if the error is related to the one mentioned in the context. However, the answer refers to error 403, not 404. Consequently, the answer is deemed incorrect.

\subsection{Deriving Heuristics from the Decision Tree}\label{sec:deriving}
The constructed decision tree represents the mental model of a human annotator. To transform this model into automated features, we need to infer the heuristics that can be extracted from the nodes in the decision tree. 

\subsubsection{Message Types}\label{sec:types}
In the decision tree, different message types follow distinct paths. For example, if a user requests an email translation, an error is unlikely due to conflicting information in the response. 

Since the type of message impacts the nature of the response, it is essential to identify the various message types. This way, we can examine how truthfulness can be assessed for each type. As far as we are aware, there is no existing classification of user message types, except those related to financial inquiries \citep{Roychowdhury-Alvarez-Moore-Krema-Gelpi-Agrawal-Rodríguez-Rodríguez-Cabrejas-Serrano-et} or chatbots for general use \citep{shah2024usinglargelanguagemodels, ji2023mitigatinghallucinationlargelanguage}. In this study, the authors utilize message type to tailor an LLM prompt to the financial objective of the question. They opted to generate question labels with particular intentions. On the other hand, our taxonomy is crafted to be more general for support bots in any field of business, not delving into the precise nature of financial questions but rather focusing on the type of information sought. This enables us to be adaptable to numerous companies and to previously unseen or unexpected questions. 

We analyzed a random sub-sample of approximately 300 messages sent to the chatbot to create the classification. These messages are message-answer pairs that were not rated, ensuring they are not subject to selection bias by the annotators \citep{SelectionBias}. Based on this analysis, we identified seven types of user messages sent by the support team. These types are derived from the decision tree and those observed in the random subset. They are:
\textit{(From now on, the underlined names will refer to these types.)}
\begin{enumerate}
    \item \underline{Error} resolution \textit{E.g., I get the error: mutation cannot be executed}
    \item \underline{Binary} answer \textit{E.g., Would it be possible to adjust tax rates manually?}
    \item \underline{Instruction} \textit{E.g., How would I adjust tax rates manually?}
    \item Cause and effect \underline{reasoning} \textit{E.g., I have adjusted tax settings, why don't I see a payslip anymore?}
    \item \underline{Action} \textit{E.g., write an email to notify customers of the new tax rates.}
    \item \underline{Unspecified} intention \textit{E.g., Good morning / I just ate a sandwich}
    \item \underline{General} information \textit{E.g., What are the tax rates in the Netherlands? / What products do you offer?}
\end{enumerate}

\begin{figure}
    \centering
    \begin{subfigure}[c]{0.45\textwidth}
        \includegraphics[width=\textwidth]{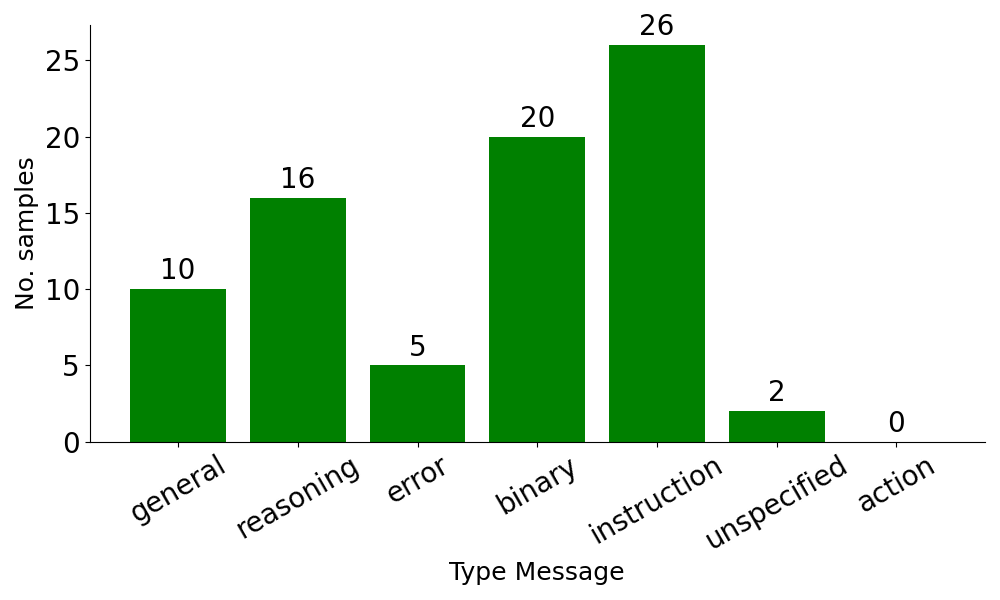}
        \caption[]{Analysis set}
    \end{subfigure}
    \hfill
    \begin{subfigure}[c]{0.45\textwidth}
        \includegraphics[width=\textwidth]{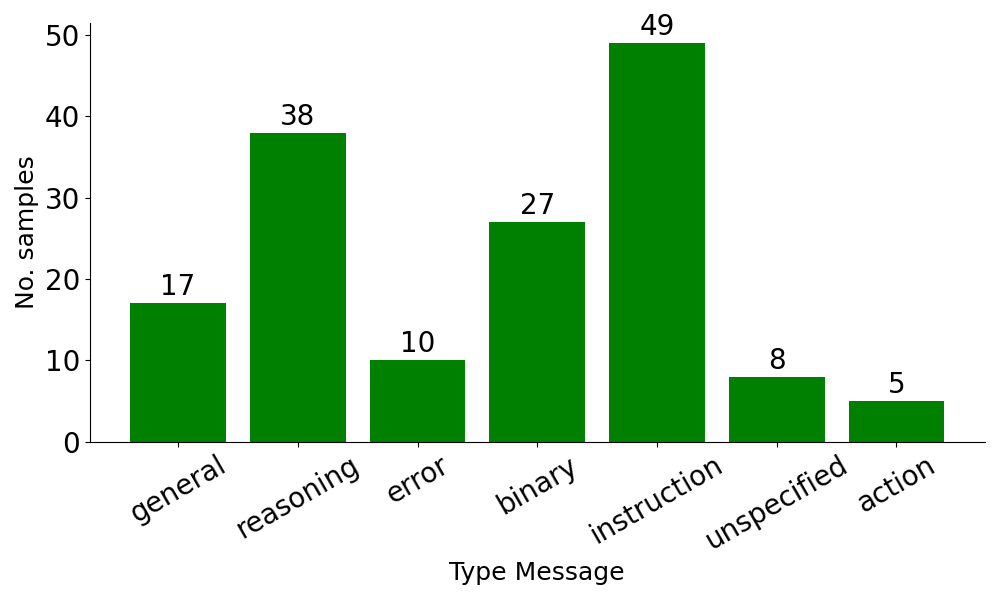}
        \caption[]{Test set}
    \end{subfigure}
    \caption{Total message-answer pairs by mistake type. The x-axis shows the message type, and the y-axis shows the number of messages per type}
    \label{fig:noMessages}
\end{figure}

Since the nature of the message influences the mistakes made, each type would require specified features to capture the mistakes relevant to their nature. Consequently, we decided to focus on a subset of message types. Initially, we only had access to the training set because there was insufficient data to create an analysis and test set. Within the training set, the types \textit{Binary} and \textit{Instruction} make up 58\% of the overall dataset, as shown in Figure \ref{fig:noMessages}. Therefore, we chose to focus on these types.
\newpage
The message types for both the analysis set and test set are labeled by the first author. The author's labeling is cross-validated by three AI developers of AFAS, who each annotate a random subset of messages, covering 77\% of the complete analysis set. The inter-annotator agreement is computed using Cohen's kappa \citep{kappaCohen}, with values below 0 indicating disagreement, above 0 agreement, and 1 perfect agreement. Kappa has been used in similar research \citep{Higashinaka_Araki_Tsukahara_Mizukami_2021, vanderLee, Lowe-Noseworthy-Serban-Angelard, Merdivan-Singh-Hanke-Kropf-Holzinger-Geist-2020}, with resulting values often ranging between 0.3 and 0.5 \citep{vanderLee}, and has been employed to evaluate the usefulness of error taxonomies for chatbots reaching a kappa of 0.44 \citep{Higashinaka_Araki_Tsukahara_Mizukami_2021}.
Our message type taxonomy achieves a Cohen's kappa of 0.65, which is considered moderate \citep{Lowe-Noseworthy-Serban-Angelard}. The author and developers have a high agreement of 0.81 for Binary and Instruction message types, with 91\% of messages consistently labeled. Given the overlap between the two types, using them together for automated scoring makes sense.

\subsubsection{Heuristics identified}\label{sec:identified}
By utilizing the decision tree and feedback from the support team, we derive the following heuristics that are syntactic and semantic in nature. 

\textit{Unspecified Components} Untrue answers may include menu items, buttons, and settings that are not specified in the context or question.

\textit{Guide Verbatim} An answer is more likely to be true if it includes a guide that is almost verbatim from the context.

\textit{Answer Contradiction} Untrue answers may include contradictions within the answer.

\textit{Error Mismatch} Untrue answers may include a mix-up of error names or codes with their solutions.

\textit{Non-appearing statements} Untrue answers may contain statements that are not present in the context, not even a variation of the statement.

\textit{Off-Context} Mistakes can be very subtle. An answer might be generally correct, but slightly off in context, such as referring to a single employee when the context is about multiple employees.

\textit{General Answer} An answer is less likely to be true if it is too general. If it is too simple or vague.

\textit{Context Synthesis} An answer is more likely to be true if multiple documents are combined to create the response.

\textit{Out of Context} The LLM might use only a small part of the context for its reasoning, meaning the exact answer may not always be clearly present, regardless of whether the answer is true or not.

\textit{Context Limitation} Some true answers receive 3 stars or fewer. While these answers are correct, a better solution exists. Although these solutions are not found in the context, they are known by a support employee.

Table \ref{tab:heuristics-per-type-message} shows the heuristics observed for each message type. This further justifies our choice to study Binary and Instruction types, given a substantial overlap of heuristics to measure truthfulness.

\begin{table*}[t]
    \centering
    \begin{tabularx}{\textwidth}{l|XXXXXX}
        \hline
        \textbf{Type Message} & {\textbf{General}} & {\textbf{Reasoning}} & \textbf{Error} & {\textbf{Binary}} & {\textbf{Instruction}} & \textbf{Unspecified} \\
        \hline
        \makecell[l]{Unspecified Components} & \ding{51} & \ding{51} & & & \ding{51} & \\
        \hline
        \makecell[l]{Guide Verbatim} & & & \ding{51} & \ding{51} & \ding{51} & \\
        \hline
        \makecell[l]{Answer Contradiction} & \ding{51} & & & \ding{51} & & \\
        \hline
        \makecell[l]{Error Mismatch} & \ding{51} & & \ding{51} & & & \\
        \hline
        \makecell[l]{Non-appearing\\Statements} & & \ding{51} & & \ding{51} & \ding{51} & \\
        \hline
        \makecell[l]{Off-Context} & & \ding{51} & & \ding{51} & \ding{51} & \\
        \hline
        \makecell[l]{General Answer} & \ding{51} & & \ding{51} & \ding{51} & \ding{51} & \ding{51} \\
        \hline
        \makecell[l]{Context Synthesis} & \ding{51} & \ding{51} & \ding{51} & \ding{51} & \ding{51} & \ding{51} \\
        \hline
        \makecell[l]{Out of Context} & \ding{51} & \ding{51} & \ding{51} & \ding{51} & \ding{51} & \ding{51} \\
        \hline
        \makecell[l]{Context Limitation} & & \ding{51} & \ding{51} & & \ding{51} & \\
        \hline
    \end{tabularx}
    \caption{Overview of heuristics for each type of message. Ranging from syntactic to semantic heuristics. x-axis showing the type of message, y-axis the heuristic}
    \label{tab:heuristics-per-type-message}
\end{table*}

\section{From Human to Automated Assessment}\label{sec:automated}
First, we explore features to predict message type. Then, we solicit features tailored to measure the truthfulness of Binary and Instruction-type messages. Finally, we curate features to generate a score. 

\subsection{Identify Type of User Message}
Due to the limited number of messages and the presence of 7 different message types, only a few samples per class are available. E.g. only 5 for type Error, see Figure \ref{fig:noMessages}. While it is common to train a machine learning model for classification \citep{Classification}, our sparse dataset is not large enough for both training and testing. Consequently, we attempt to develop a rule-based algorithm, based on human analysis, to identify the types of messages. 

Every message type is characterized by its structure, with distinct patterns arising according to the type. For each of the types, a list of common patterns observed in the analysis set is created. This does not apply to the type Action, as no message of that type is present in the analysis set. 

In natural language processing it is common to employ pre-processing before the actual text processing, such as classification. Examples of these techniques include stopword removal \citep{StopwordsRemoveObie, StopwordsRemoveCarrebi, StopwordsRemoveZhan}, lemmatization \citep{LemmatizePhong}, punctuation removal \citep{StopwordsRemoveObie, StopwordsRemoveZhan, PuncuationTahvili} and lowercasing \citep{StopwordsRemoveObie, StopwordsRemoveCarrebi, StopwordsRemoveZhan, PuncuationTahvili}. These methods help overcome minor variations between identical words or sentences, reducing noise in the data. For the message type prediction, the text is lowercased as the lists with patterns are lowercased and not case-sensitive.

For a message, our approach scans for words or patterns from the first list. If a match is found, a message is assigned to the corresponding message type. If no match is found, the system checks for a match in the next list. This is continued till a match is found and the type unspecified is assigned otherwise. Therefore, if a question contains words from multiple lists, it is assigned the type of the first list with which it shares a word. The order of the lists is determined based on their performance on the analysis set, focusing especially on the accuracy of classifying Binary and Instruction types. This is because the main interest is in accurately classifying these two types, as only those types will be scored. This leads to the following sequence of lists; for complete lists, please refer to the replication package \citep{Replication}:

\begin{itemize}
    \item[] error : \textit{{['error', ...]}}
    \item[] general : \textit{{[' explanation', 'what is', ...]}}
    \item[] reasoning : \textit{{['why', 'how can this', 'why', ...]} {or has no question mark}}
    \item[] instruction : \textit{{['how ', 'where ', ...]}}
    \item[] binary : \textit{{['possible', 'can ', ...]}}
    \item[] unspecified : \textit{if there are no matches in the lists above}  
\end{itemize}

In the automated scoring process, the message type is first predicted. If it is of either Binary or Instruction type, a score is assigned. If another type is predicted, no score is assigned, as other types are not within the scope of our automated metric.

\subsection{Feature Selection}\label{sec:features}
With truthfulness defined and the heuristics for a true answer established, the next step is transforming heuristics into automated features. Note that LLM not having the knowledge about the product is a key factor in this research. As the external trained model has no knowledge about the company, all info mentioned should be present in the context, otherwise it is hallucinating \citep{Roychowdhury-Alvarez-Moore-Krema-Gelpi-Agrawal-Rodríguez-Rodríguez-Cabrejas-Serrano-et}. Therefore, we add verifying whether the content is contained within the context to the heuristics. At the beginning of our study, we made the decision to avoid using LLMs for answer evaluation. This decision was based on existing evaluations with LLMs, which have proven to be expensive and time-consuming \citep{lin2023llmevalunifiedmultidimensionalautomatic}. Our goal is to provide real-time scoring so that users do not have to wait for an additional model call. Additionally, LLMs are inconsistent in scoring and error-prone, often producing different outcomes depending on the prompt and model used \citep{lin2023llmevalunifiedmultidimensionalautomatic}. We also aim for consistent scoring for each question-answer pair to ensure reliable comparisons.

To translate the heuristics to automated features, we use literature to find features to automate this. To find relevant papers, the words from the following non-exhaustive list are combined: \textit{metric, correctness, score, evaluate, chatbot, conversational bot, conversational agent, hallucination, characteristics, wrong, correct, NLP, education, grading, measure, automated}. Using the word combinations, papers are found utilizing various scientific databanks including ResearchGate\footnote{https://www.researchgate.net/}, Google Scholar\footnote{https://scholar.google.com/}, and IEEE Xplore\footnote{https://ieeexplore.ieee.org/Xplore/home.jsp}. Additionally, relevant papers are used for snowballing, both backward and forward. About 70 relevant papers are identified, which either define a non-automated metric for chatbots or an automatic metric for general text grading.
We identified two relevant fields with related features: automated metrics for chatbots and automated grading of student answers. A full list is constructed with potential relevant features, see the replication package \citep{Replication} for all identified features. 

The full list is compared to the heuristics, if a feature cannot replace any heuristic due to insufficient overlap in working, it is excluded. Next, we implement these filtered features, features inspired on literature and features for heuristics that are not covered by any existing feature.

Each of the implemented features is assessed for Spearman correlation \citep{Spearman} with human evaluation. If a feature has a positive correlation and a p-value lower than 0.10, it is selected for the final selection. While a correlation of 0.10 is not statistically significant it is deemed sufficient, as the actual selection will take place during the final selection. If a feature doesn't meet the correlation criteria but has demonstrated effectiveness in relevant literature, it is tested if it distinguishes between true and untrue answers. If it can differentiate at least some of these answers, it is selected for the final selection. Additionally, in the final selection, features that perform the exact same function as the heuristic they are intended to replace, such as verifying the presence of a word, are included.

After this raw selection of most promising features, the final selection is done using an ablation study \citep{AbblationThesis, AbblationNN}. We assess whether removing the feature impacts the significant correlation between the combined features and the human evaluation for the analysis set. If the removal decreases this correlation, we keep the feature. This hierarchy of filtering steps enable an initial rough selection with limited confidence, as features are ultimately chosen only if they pass the strictest selection criterion, which is the final step.

The accepted features are listed, indicating whether they were selected in the initial filtering due to correlation (\correlation), distinguishing some answers in combination with literature (\literature), or directly replacing a heuristic (\heuristic). Additionally, it is noted whether the feature can handle text translated into English (\googletranslate).

\textbf{\correlation Company-Specific Terms} - This feature is developed based on the General Answer heuristic. The assumption is that an answer should not include general terms when describing a solution. The 10,000 most frequent used words in AFAS its help documentation are checked against all words in the Dutch Wikibooks dataset \citep{dhruvil_dave_2021}. Words not found in this dataset are considered company-specific, and answers with such words are deemed more truthful.

\textbf{\heuristic Components Defined} - As per the Unspecified Component heuristic, an answer should only include existing components. Which is influenced by the feature introduced by \cite{Roychowdhury-Alvarez-Moore-Krema-Gelpi-Agrawal-Rodríguez-Rodríguez-Cabrejas-Serrano-et}, where they verify the precise financial numbers within the context. In the help documentation, components are defined structurally, and the LLM preserves this structure in its responses, even if mentioning non-existing components. Therefore, components can be extracted using REGEX. If an answer includes components not defined in the context, it is less likely to be true.

\textbf{\correlation \literature Complex Answer} - Following the General Answer heuristic, we check for the existence of words indicating a complex text, as introduced by \cite{Kumar-Aggarwal-Mahata-Shah-Kumaraguru-Zimmermann-2019}. While no such lists are available in scientific literature for Dutch, we created four lists based on signal words from Genootschap Onze Taal\footnote{\textit{(Society Our Language):} https://onzetaal.nl/taalloket/signaalwoorden-lijst (29-05-2024)} and Boom NT2\footnote{https://www.nt2.nl/documenten/luisteren\_op\_b2/overzicht\_van\_ signaalwoorden.pdf (29-05-2024)}, each corresponding to different types of complexity: perspective, comparison, examples, and reasoning. Full lists can be found in the replication package \citep{Replication}. The presence of words from various lists increases the likelihood that the answer is true.

\textbf{\literature Prompt Overlap} - \cite{Roychowdhury-Alvarez-Moore-Krema-Gelpi-Agrawal-Rodríguez-Rodríguez-Cabrejas-Serrano-et} introduced the Prompt Uniqueness feature, which indicates that answers are less likely to be good if they repeat parts of the question. In contrast, \cite{Kumar-Aggarwal-Mahata-Shah-Kumaraguru-Zimmermann-2019} proposed Prompt Overlap, suggesting that some overlap is expected and, when present, indicates a better answer.

We examined both and discovered that a prompt overlap suggests a higher likelihood of encountering a true answer.

\textbf{\correlation \literature \googletranslate HAL} - In order to address the Off-Context heuristic, the HAL technique from the study by \cite{Lund_Burgess_1995} is implemented. This feature measures how often pairs of words appear together within a sliding window of varying sizes, giving higher scores to pairs that frequently occur next to each other. The intuition is that if a word or setting frequently appears together in the answer, it should also be close together in the context. If not, it is less likely a true answer.

\textbf{\correlation \googletranslate Subject Combination} - This feature is founded on two heuristics: Non-appearing statements and Off-Context. For both answer and context, relationships between verbs and nominal subjects are identified by extracting pairs connected by a subject \textit{('n\_subj')} dependency using the spaCy dependency parser\footnote{https://spacy.io/usage/linguistic-features/\#dependency-parse}. If each pair in the answer is present in the context, it is more likely true. 

\textbf{\correlation \googletranslate Verbatim Guide Defined} - This feature is created to capture the heuristic Guide Verbatim. Since guides follow a fixed structure in help documents, and this structure is adopted by LLM responses, they can be extracted using REGEX. Steps for each guide are first extracted from both the answer and context. Then, it verifies if a similar guide exists in the context by comparing the guide lengths and using cosine similarity with spaCy\footnote{https://spacy.io/} to assess the similarity of all steps. If an extremely similar guide exists in the context, the answer is more likely to be true.

\subsection{Features to Score}
To go from these features to a score, all features are normalized between 0 and 1. Following the method of \cite{Roychowdhury-Alvarez-Moore-Krema-Gelpi-Agrawal-Rodríguez-Rodríguez-Cabrejas-Serrano-et}, we sum the features together. We normalize the sum between 1 and 5, adhering to the rating of human annotators.

\begin{figure}
    \centering
    \includegraphics[width=0.8\linewidth]{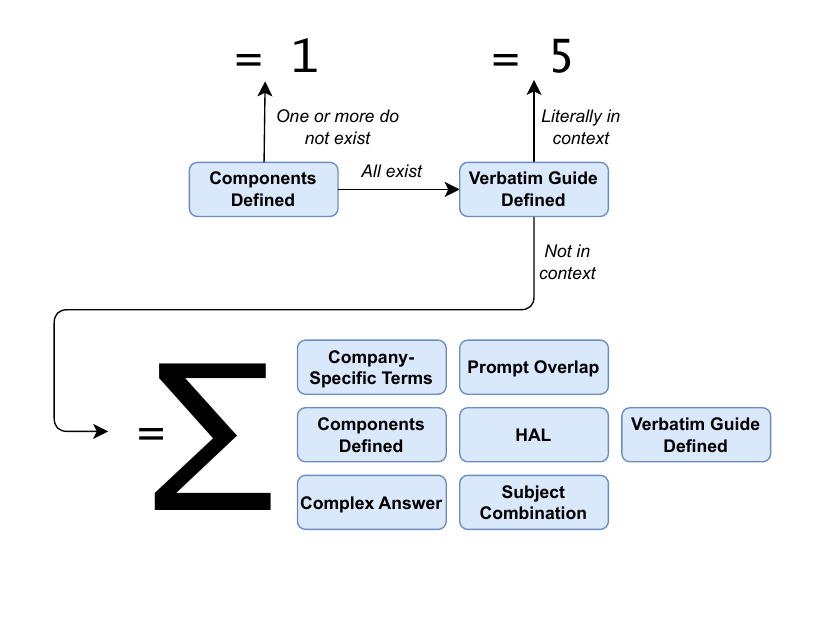}
    \caption{Illustration of the scoring process, where scores range from 1 to 5. A score of 1 is assigned if any component is missing, and 5 if a guide in the answer matches one in the context. Intermediate scores are obtained by summing the outputs of the other metrics}
    \label{fig:scoreVsSum}
\end{figure}

In this approach, each feature is given equal weight. However, the decision tree reveals that an answer is untrue anyway if it contains non-existing components, leading us to define a second score where answers with such components receive a 1-star rating. In addition, the analysis teaches that if a message contains a verbatim guide, the answer is anyway true. Therefore, if an answer contains such a guide, it is rated with 5. This is visually explained in Figure \ref{fig:scoreVsSum}.

\section{Evaluation}\label{sec:evaluation}
Our test set is obtained from the feedback system a few weeks after the extraction of the analysis set. As the bot is developed during that time, the answers given by the bot also changed.

\subsection{Method of Evaluation}
We evaluate both message type prediction and score prediction. The prediction is measured using F1 score, precision, recall, and accuracy, which show how often the label is predicted correctly. The main goal is to label messages as Binary or Instruction types, as only these will be scored. Therefore, our final evaluation emphasizes how well it predicts these two types.

Evaluation on the scoring is done with the Dutch text and English text. For the Dutch text, tests showed that the features worked best when lemmatized and lowercased. In section \ref{sec:features} the \googletranslate symbol shows for which features English text is used in the English scoring version. These features utilize externally trained packages, which work with English text as well. The other features still use Dutch text because they rely on custom words and regex fine-tuned for Dutch. The English texts show the best result when lemmatized, lowercased and stripped from stopwords.

The evaluation of the resulting scores is conducted across three gradations: overall performance, variations in prediction, and situations where the approach is ineffective. For the overall performance, the Spearman correlation between the prediction and the human evaluation is used. This is commonly used in similar research \citep{Pang-Nijkamp-Han-Zhou-Liu-Tu-2020, Tao-Mou-Zhao-Yan-2018, Haque_Eberhart_Bansal_McMillan_2022}. With significance levels of 0.05 \citep{Huang-Ye-Qin-Lin-Liang-2020, Lowe-Noseworthy-Serban-Angelard} and 0.01 \citep{Guan_Huang_2020, Tao-Mou-Zhao-Yan-2018}. Secondly, we assess the deviations using an error margin. As the rating of the human annotators is discrete, and our metric continuous. An error margin of 1 is used, as it reflects whether a positive rated message ($>3$) is rated above three by our metric, and vice versa. Finally, the scenarios for which our metric does not work are determined by manually assessing mistakes made.

\subsection{Results}

\subsubsection{Message Type Prediction}
Table \ref{tab:resultsMessageType} shows the evaluation of the type prediction for both, analysis and test set. The overall performance is shown, along with the predictions for the types Binary and Instruction. All patterns utilized in the prediction are derived solely from the analysis set. This may result in many messages in the test set not matching any pattern; however, only 18\% of the test set is labeled as unspecified, suggesting that at most 18\% of the messages contain unseen patterns. Among the messages labeled as Binary and Instruction, only six messages do not match any pattern.

Bear in mind that the prediction is solely based on rule-based checks. After all, we achieve an F1 score of 0.77 on the test set for type Binary and Instruction, and an F1 score of 0.80 for Instruction type. Table \ref{tab:resultsMessageType} displays that 82\% of the messages that will be scored are genuinely of type Binary or Instruction. Additionally, around 72\% of messages identified as Binary and Instruction are accurately predicted as such, and consequently scored, as indicated in Table \ref{tab:resultsMessageType}.

\begin{table*}[htbp]
    \centering
    \begin{tabularx}{\textwidth}{Xcccc}
        \hline
         & \bf F1 Score & \bf Accuracy & \bf Precision & \bf Recall \\
        \hline
        Analysis (all) & \textbf{0.81} & 0.80 & 0.83 & 0.80 \\
        Analysis (Binary) & 0.95 & 0.95 & 0.95 & 0.95 \\
        Analysis (Instruction) & 0.96 & 0.96 & 1.00 & 0.92 \\
        Analysis (is Binary or Instruction) & \textbf{0.96} & 0.96 & 0.98 & 0.93 \\
        \hline
        Test (all) & \textbf{0.62} & 0.61 & 0.65 & 0.61 \\
        Test (Binary) & 0.62 & 0.62 & 0.61 & 0.63 \\
        Test (Instruction) & 0.80 & 0.81 & 0.90 & 0.71 \\
        Test (is Binary or Instruction) & \textbf{0.77} & 0.77 & \textbf{0.82} & \textbf{0.72} \\
        \hline
    \end{tabularx}
    \caption{Automated message detection performance on training and testing datasets: \textit{all} for overall classification, \textit{Binary} and \textit{Instruction} for labeling respective types, and \textit{is Binary or Instruction} for distinguishing either type or neither}
    \label{tab:resultsMessageType}
\end{table*}

\subsubsection{Overall}
The overall performance indicates that the Dutch features have been tailored and optimized for the analysis set, as they exhibit better performance on this set compared to the English features and the Test set, as shown in Table \ref{tab:finalResults}. However, this is not the case for the translated version, where the test set outperforms the analysis set. All of the features are fine-tuned on the Dutch data, therefore it is possibly overfitted. For the test set, it concludes that the translation has a positive effect on the prediction. This indicates that the external packages used work better with English than with Dutch text.

To contextualize the correlation of our metric, we refer to the most comparable metric in terms of definition discovered in the literature, developed by \cite{Mehri-Eskenazi-2020}. They evaluate answers based on correctness and achieve a correlation of 0.13. Notably, the Dutch and English versions exceed this result, with correlations of 0.28 and 0.37, respectively.

\begin{table}[!hbtp]
    \centering
    \begin{tabular}{cl}
        \hline
        \bf Set & \bf Score \\
        \hline
        Analysis & 0.45** \\
        Test & 0.28* \\
        Analysis \footnotesize{(Translated)} & 0.30* \\
        Test \footnotesize{(Translated)} & 0.37** \\
        \hline\\
    \end{tabular}
    \caption{Spearman Correlation between Human truthfulness evaluation and automated Score results. * for p$<$0.05, ** for p$<$0.01}
    \label{tab:finalResults}
\end{table}

\subsubsection{Deviations}
To assess the accuracy and deviation, see Table \ref{tab:actualScores}. It performs especially well in rating the high and low rated messages, not the neutral. Additionally, Table \ref{tab:actualScores} shows that over half of those rated with 1 star are containing non-existing components. Notably, 60\% of neutral responses include a non-existent component. This may imply that the answer is true and comprehensible. However, due to the incorrect terminology used, the answer cannot be considered entirely true. Table \ref{tab:actualScores} shows a threshold of 3, which indicates neutrality. Messages with a score above 3 should be considered true, and those below 3 should be considered untrue. As demonstrated, 67\% of the 1-star rated messages have a score below three, while only 21\% of the 5-star rated messages do. Therefore, by not sending any messages with a score lower than 3 to the user, 64\% of the messages do not need to be judged by hand by the support team, and only 21\% of the 5-star rated messages will be discarded.

\begin{table*}[htbp]
\begin{tabular}{lccccc}
    \hline
    & \textbf{Rated 1} & \textbf{Rated 2} & \textbf{Rated 3} & \textbf{Rated 4} & \textbf{Rated 5} \\
    \hline
    \bf Margin 1 & 55\% & 33\% & 20\% & 67\% & 40\% \\
    \bf Score $==$ 1 & 55\% & 0\% & 60\% & 0\% & 12\% \\
    \bf Score $<$ 3 & 64\% & 67\% & 60\% & 33\% & 21\% \\
    \bf Score $>$ 3 & 36\% & 33\% & 40\% & 67\% & 79\% \\
    \bf Score $==$ 5 & 18\% & 0\% & 20\% & 50\% & 26\% \\
    \hline
\end{tabular}
    \caption{Automated answer rating accuracy ($\pm1$ error margin) and percentage of what score is predicted (below/above neutral) per actual label. These conditions show how often negative-rated messages are correctly predicted as such, and positive as positive. Score 1: failure on Components Defined, Score 5: success on Verbatim Guide Defined}
    \label{tab:actualScores}
\end{table*}

\subsubsection{Scenarios not working}
To identify the mistakes made by the chatbot and guide future work, scores that deviate much from the ground truth are analyzed. First, a downside of using REGEX is that it can miss some components or extract unrelated text. E.g\textit{"The salary button should be clicked" Instead of expected: "Click on: salary"}. Another recurring problem is the ambiguity when detecting the heuristic Off-Context. Generally, an answer may be correct, but not for a specific exception. E.g. \textit{User asks a question about a Nurse organization, however for these organization different laws hold true.} Third, relatedness and completeness influence the rating alongside truthfulness, even though the ranking is distinct for each dimension. Lastly, sometimes the LLM hallucinates correctly, when it is based on sparse information in the context. E.g. \textit{Q: "What if I click the salary button" A:"It shows a salary overview". Although not explicitly stated in the context, the name of the button effectively inspires this correct hallucination.}

\section{Threats to Validity}
\textbf{Internal Validity} includes threats related to the methods and processes of the study. First, at AFAS, as we developed the metric, the chatbot also evolved. The implication is that our data changes over time, and our analysis set differs from the test set. We do not see this as a problem but believe that this ensures our results are transferable across the evolving chatbot configurations. Second, our approach hinges on contextual information derived from help documents. Any missing, outdated, or incorrect information relating to the context implies incorrect validation in practice. This threat is hard to mitigate, but it will make keeping documentation up-to-date and complete increasingly important for such systems to work. Further, our messages are annotated only by a support employee working on them. As a result, differences in perspectives will be reflected in rating. That being said, since our annotators are topic experts, they likely spot some if not all, mistakes. 

\textbf{External Validity} includes threats to generalizability. We propose metrics for two message types and derive insights from AFAS. The bigger question is, who can use this work? Unlike prior work, our metrics are not linked to the financial sector and can be easily adapted to other fields. Replicating our work can be the first step in seeing feasibility. And even if our metrics do not apply to a environment, the methodology can inspire the identification of custom metrics. While the usability of the metric has not yet been tested by others in real-world settings, we evaluated our recommendations on a manually annotated training set, reflecting how the support team perceives our recommendations. In future, we can add our recommendations to the other chatbots to see how it works in practice.

\section{Lessons Learned}
The customer support team is on a challenging mission to accurately and efficiently respond to all kinds of customer queries. One of the most persistent challenges we faced was defining what a right answer means. We found that all these valid responses share three requirements: \emph{truthfulness, relatedness, and completeness}. This definition is applicable to any support chatbot, as it uses criteria that are universal applicable across various domains and languages. These requirements provided us with a deeper understanding of the chatbot its limitations. Utilizing the definition of correctness and its requirements allows for a meaningful evaluation, whether conducted by humans or automated systems.

\subsection{Scope your metric}
At the outset of our research, we aimed to develop a metric capable of identifying correct answers for all incoming user questions. However, we quickly realized that it would be hard to achieve a generic metric. We discovered that not all types of questions can be treated the same. A notable observation of this study is that the type of user message influences the nature of mistakes. Initial signs of this observation in the literature \citep{Roychowdhury-Alvarez-Moore-Krema-Gelpi-Agrawal-Rodríguez-Rodríguez-Cabrejas-Serrano-et}, which gives the impression that this observation extends to other chatbots. By identifying the different types, we were able to use targeted heuristics instead of applying the same approach to everything. By narrowing the scope of our evaluation, we move away from our original goal, but we gain practical impact. \textit{\textbf{It can be advised to tailor the metrics to the question types, a general evaluation does not work.}}

\subsection{Do not aim for perfection, but aim for impact}
While our approach is not without its flaws, AFAS has integrated our solution into their system and is testing whether some answers can be directly sent to the user. Table \ref{tab:actualScores} shows that, among the 5-rated messages, our system successfully identified 26\% as such. For the wrongly generated answers, we were able to detect 55\%. Considering the annual number of inquiries (Section \ref{sec:industrialsetting}) and the fact that roughly half of all messages are either binary or instructional in nature, we estimate that detecting even 26\% of the 5-star rated messages could save approximately 15,000 hours per year. This reduction in workload would allow the support staff to focus on more complex inquiries and provide users with near real-time responses.

Beyond effeciency gains, the research led to unexpected side effects. By learning from the common mistakes of the chatbot, the support team at AFAS revisited the reformulation of questions to obtain the correct answers from the chatbot. An other side effect arose in the documentation flow, since the chatbot uses the available help documentation as context, the quality of the documentation is directly related to the quality of the chatbot. A wrong answer can be caused by bugs in the chatbot, but can also be caused by mistakes in the documentation. In that sense, the chatbot is similar to a new colleague onboarding who misses documentation or reads flawed documentation \citep{TugOfWarOfAids, CamerasOn}. 

This emphasizes that automation and the integration of chatbots into daily workflows have a widespread impact throughout the entire process. Analyzing these chatbots can uncover systemic weaknesses within a company, such as flaws in documentation. It illustrates how automation efforts can affect broader team practices.

Analyzing chatbot mistakes and the feedback loop, helps the support team by refining their process and question crafting. In addition, We were able to exploit the feedback given on chatbot answers to improve the quality of the documentation. By constructing a feedback loop from the feedback given on chatbot answers to the documentation team we can identity parts of the help documentation that are of low quality and are most useful to improve. The chatbot serves as a tool for identifying parts of the documentation that need attention\citep{DocsPractitionersPerspective}.

\textit{\textbf{Organizations should be mindful of what they are evaluating. Are they evaluating the correctness of the bot, or the documentation. In addition, by integrating error analysis into support workflows and documentation review, teams can enhance both the quality of responses and the underlying knowledge base.  A thorough evaluation of your chatbot can lead to improvements that extend beyond the chatbot itself.}}

\subsection{Do not underestimate the power of custom features}
Since we are working with a flexible software product that offers many features, building a dataset is challenging. It requires input from support team experts. In addition, we were dealing with specific company knowledge and nuances, which are difficult to capture using generic tools.

To address this, we experimented with custom features. We showed that by using targeted checks and leveraging AFAS its standardized documenten style, we could effectively distinguish between correct and incorrect responses. This tailored approach allows us to validate even nuances specific to the company domain. Leveraging company documentation as ground truth, serves as a powerful validator when reference answers are not available.

This approach of custom features can be leveraged if a chatbot is operating in a specialized domain knowledge or requires company specific nuances to be taken into account. Our metrics can be adapted by learning from the documentation style of a particular company, other metrics focus on textual comparison which are independent of language or business which make them even more adaptable.

\textit{\textbf{We discovered that even simple targeted heuristics and well-structured documentation can improve response validation. Organizations should treat company documentation as a powerful ground truth.}}

However, our approach also has limitations in detecting mistakes. Some of the mistakes are due to the limitations of automated features, while others are hard even for the human annotator to detect. We identified three edge cases that are difficult for human annotators to detect and require a thorough knowledge of the company. These cases include situations where a superior solution exists, solutions with undesirable side effects, and the use of terms that change meaning when used in the context of AFAS. Capturing these nuances using automated features is even harder.

To address these limitations, a direction is to construct a knowledge base that learns from all incoming messages and their answers. This can supplement the information in the help documentation and help detect mistakes due to overseen side effects or from out-of-context. Likewise, we can develop a neural network that learns from our system to capture the subtle nuances our approach misses. Prior work on this topic has shown promising results \citep{Tao-Mou-Zhao-Yan-2018, Lowe-Noseworthy-Serban-Angelard, Singh-Suraksha-Nirmala-2021, Yan-Song-Wu-2016, Huang-Ye-Qin-Lin-Liang-2020, Pang-Nijkamp-Han-Zhou-Liu-Tu-2020} and student answers \citep{Xue-Tang-Zheng-2021, Ormerod-Lottridge-Harris-Patel-vanWamelen-Kodeswaran-Woolf-Young-2022}. This way, we graduate from a workable solution to a scalable solution.

\subsection{Translate your data}
Compared to most studies in English, assessing correctness in Dutch was challenging. This was evident in the performance of results when we used Dutch text versus English text; there was a performance gain with the English text. This challenge is relevant to many other regional languages, that are less studied than large languages like English. Since the most widely used software packages for natural language processing are in English. For example, the Python software packages used in this work are tailored for English text. When working in languages with weaker NLP support, translation into English can be a practical workaround to leverage stronger tools and models. \textit{\textbf{We extend this recommendation to other similar explorations in regional languages to consider translating the text to English for performance gain}.}

\subsection{Structured Approach}
In this study, we evaluated the truthfulness of responses generated by a Dutch-support chatbot using simple and adaptable metrics. Our methodology translates the human decision-making process into heuristics and then into measurable metrics, providing a structured approach to assess correctness.

Since this approach is based on how evaluators naturally judge responses, it is not specific to AFAS or limited to Dutch. Any organization can employ the same process to evaluate correctness within their own domain. The value lies not only in the specific metrics but also in the methodology itself, which involves employees in defining correctness criteria and translating these into metrics.

\textit{\textbf{Other organizations can replicate this process by capturing the decision-making process of their end-users and converting it into automated metrics. This makes correctness assessment both structured and adaptable to various domains and languages.}}

\section{Related Work}
As our research assess the correctness of content, two relevant fields are identified: Natural Language Generation (NLG) and Automated Answer Grading. The chatbot in question will be categorized as user-driven \citep{10.1007/978-3-030-17705-8_13} and support bot, specifically a Generative Question Answering (GQA) bot \citep{Ji_2023}.

\subsection{Natural Language Generation}
In previous studies of NLG, human evaluation is often used \citep{adiwardana2020humanlike, deriu2020spot, jamanetworkopen.2023.36483, Serban_Sordoni_Lowe_Charlin_Pineau_Courville_Bengio_2017}. To date, several studies have investigated how human evaluation can be automated. Focusing on dialogue quality \citep{Yeh-Eskenazi-Mehri-2021, Huang-Ye-Qin-Lin-Liang-2020, Pang-Nijkamp-Han-Zhou-Liu-Tu-2020}, or on a single message-answer pair. Our study falls under the latter, known as turn-level metrics.
Much of the literature on turn-level metrics is focused on comparing text embeddings \citep{Merdivan-Singh-Hanke-Kropf-Holzinger-Geist-2020, zhang2020bertscore, banerjee2023benchmarking, Das-Verma-2020}. It either tests relevancy \citep{Merdivan-Singh-Hanke-Kropf-Holzinger-Geist-2020, zhang2020bertscore} or improvement \citep{banerjee2023benchmarking, Das-Verma-2020}. Improvement is measured along linguistic features like readability, syntactic style and complexity \citep{Das-Verma-2020}. There are relatively few studies in the area of content-specific aspects. The existing research is focused on rational answers \citep{Phy-Zhao-Aizawa-2020, Tao-Mou-Zhao-Yan-2018} and helpful answers \citep{Gupta-Rajasekar-Patel-Kulkarni-Sunell-Kim-Ganapathy-Trivedi-2022} rather than on correctness of answers. \cite{Ji_2023} mention the concept of correctness and highlight that in the field of GQA, there is an absence of standardized definitions. They also note that current methods for assessing the factual correctness of answers usually rely on human evaluation, and better automatic evaluation is needed.

\citep{Roychowdhury-Alvarez-Moore-Krema-Gelpi-Agrawal-Rodríguez-Rodríguez-Cabrejas-Serrano-et} suggest a framework for a financial bot that includes confidence monitoring. This monitoring aims to identify whether the LLM is hallucinating in comparison to the context, with a focus on numerical hallucinations. While we have drawn inspiration from some features to evaluate hallucinations, our approach assesses content correctness using an expanded definition of truthfulness. Further, while they propose features for safeguarding a financial bot for decision makers, our features are focused on a metric for chatbots designed for more general content and support questions. This metric is assessed by comparing it to human evaluations. 

While hallucination is well researched \citep{SurveyHallucination} and overlaps with truthfulness, they are not identical. Both involve producing information not present in the document base. However, hallucinated information can sometimes be accurate (e.g., correct or extrinsic hallucination \citep{Ji_2023}.) On the other hand, non-hallucinated information is not always accurate. E.g., if non-hallucinated information is present in a different context than in the help documentation. 

Correctness is mentioned before in literature as a metric, in the research from \cite{Mehri-Eskenazi-2020}. They use a LLM to rate generated answers on various dimensions, among which, correctness. Using an LLM in automated evaluation effectively assesses coherence and consistency \citep{Ke_Zhou_Lin_Li_Zhou_Zhu_Huang_2022}. However, this method has been found to be costly and inconsistent \citep{lin2023llmevalunifiedmultidimensionalautomatic}. To address these issues, we explored the use of deterministic methods.

\subsection{Automated Answer Grading}
There are a number of similarites between the field of automated metrics and automated grading in education. Where our metric grades the answer of a chatbot, these models and features predict the grade a teacher would give the answer of a student \citep{Jamil-Hameed-2023, Kumar-Aggarwal-Mahata-Shah-Kumaraguru-Zimmermann-2019, Mukti-Alfarozi-Kusumawardani-2023, Ormerod-Lottridge-Harris-Patel-vanWamelen-Kodeswaran-Woolf-Young-2022, P-B-2022, roy2016iterative, Vij-Tayal-Jain-2019}. \cite{Kumar-Aggarwal-Mahata-Shah-Kumaraguru-Zimmermann-2019} discusses numerous features used in both automatic essay grading and short answer grading. Some of these features, as well as those introduced by \cite{roy2016iterative}, are either directly or indirectly incorporated into our research. However, their features are designed for training a model that is based on multiple correct answers for a single question. This approach is not directly applicable to automated metrics without using reference answers.

\section{Conclusions}
To improve customer experience and enable the support team to answer customer queries faster, we embarked on a journey to assess the correctness of answers generated by Dutch support chatbot AFAS. The support team at AFAS played a crucial role in this process, especially considering the complexity of our task - the text was in Dutch, and we had sparse data for training, meaning rated answers were only limited available. We proposed metrics inferred from how the support team at AFAS assesses correctness. These metrics look at user messages, help documentation at AFAS to infer correctness, and are generic to assess unseen situations. 

Our results inspires how the support team queries the chatbot. Furthermore, we estimate a gain of up to 15,000 hours annually through accurately identifying incorrect generated answers. Our study also offers recommendations to chatbot-building software companies. With this approach they can improve the quality of their chatbot by implementing guardrails that prevent the bot from telling lies. These faulty answers can be used as trainings data to improve the answers, or can be simply marked and not be returned to the customer. We have focused on the correctness of answers and our approach thus should be extended to other characteristics as well. Still we believe that this study serves as inspiration for comparable chatbot systems.

The proposed definition of correctness provides a structured basis for human feedback. The three dimensions truthfulness, relatedness and completeness enable deeper insights into the shortcomings of a chatbot.

When aiming for an automated metric, our methodology can be adopted, which is not restricted to a particular domain. Take the automotive industry, a mechanic and a salesman may both query a chatbot about the same car. Yet, their need of information will differ and thus their evaluation. The decision tree helps capture their way of thinking, which can then guide the design of heuristics and metrics.

The metrics we suggest can be used as a solid first step for building quick, custom measures, while the overall method gives a practical foundation for anyone who wants to design automated metrics that fit their own chatbot.

\newpage
\subsection*{Contributions}
All authors have contributed to the conception and design of the study. The study originates from the MSc Thesis of Herman Lassche. He performed material preparation, data collection, and analysis, with major contributions from Michiel Overeem and Ayushi Rastogi. Michiel Overeem was mainly involved in material preparation and data collection, while Ayushi Rastogi contributed primarily to the analysis. Both also assisted in other parts of the process. Herman Lassche wrote the initial draft of the manuscript, and all authors provided feedback during iterations on this draft. Every author has read and approved the final manuscript.

\subsection*{Data and Materials Availability}
The data contains sensitive information about AFAS's customers and AFAS itself. Therefore, it cannot be shared. Anonymizing the data is not possible, as it would still allow processes, questions, etc., to be traced back to the customer. This makes replication more challenging; hence, we have included a dummy data file in our \cite{Replication}. All code used in the study is available in our replication package. In addition, more comprehensive versions of the tables can be found in that package, highlighting more features examined as investigated in the thesis.

\subsection*{Usage of AI Tools}
During the preparation of this work the authors used ChatGPT (mainly GPT-4o) in order to rephrase sentences and check for language mistakes. After using this tool/service, the authors reviewed and edited the content as needed and takes full responsibility for the content of the publication.

\subsection*{Funding}
During the research, Herman Lassche received an internship fee from AFAS. After the research, he became employed at AFAS and receives a salary. Michiel Overeem also receives a salary from AFAS. All needed materials and licenses are funded by AFAS. Ayushi Rastogi has no financial links to AFAS. No additional funding is involved.


\newpage

\end{document}